\documentclass[letterpaper, 10 pt, conference]{ieeeconf}  

\usepackage{cite}
\usepackage{amsmath,amssymb,amsfonts}
\usepackage{algorithmicx}
\usepackage{algpseudocode}
\usepackage{algorithm}
\usepackage{graphicx}
\usepackage{textcomp}
\usepackage{xcolor}
\usepackage{afterpage}
\usepackage{placeins}
\usepackage{booktabs}
\usepackage{array}
\usepackage{hyperref}
\usepackage{color, colortbl}
\definecolor{taupegray}{rgb}{0.55, 0.52, 0.54}
\definecolor{gainsboro}{rgb}{0.86, 0.86, 0.86}

\IEEEoverridecommandlockouts                              

\def\BibTeX{{\rm B\kern-.05em{\sc i\kern-.025em b}\kern-.08em
    T\kern-.1667em\lower.7ex\hbox{E}\kern-.125emX}}

\newcommand{\thingname}[0]{demonstration sidetracks}
\newcommand{\ThingName}[0]{Demonstration Sidetracks}

\title{\LARGE \bf \ThingName{}: Categorizing Systematic Non-Optimality in Human Demonstrations}
\author{Shijie Fang$^{1,\dagger}$, Hang Yu$^{1,\dagger}$, Qidi Fang$^{1}$, Reuben M. Aronson$^{1}$, Elaine S. Short$^{1}$ 
\thanks{\textsuperscript{†}These authors contributed equally to this work.}
\thanks{$^{1}$Tufts University School of Engineering, Computer Science. Medford, Massachusetts, United States of America
        {\tt\small \{shijie.fang, hang.yu625917, qidi.fang, reuben.aronson, elaine.short\}@tufts.edu}}%
}

\begin{document}
%

%
%

\maketitle              

\begin{abstract}
Learning from Demonstration (LfD) has become a popular approach for robots to learn new skills, despite most LfD methods suffering from imperfections in human demonstrations.
Prior work in LfD often characterizes the sub-optimalities in human demonstrations as random noise.
In this paper, we explored non-optimal behaviors in non-expert demonstrations and showed that these behaviors are not random and have systematic patterns: they form systematic \thingname{}. 
We used a public space study dataset from our previous work with 40 participants and a long-horizon robot task.
We recreated the experimental setup in a simulation and annotated all the demonstrations. 
We identified four types of \thingname{}, \textit{Exploration}, \textit{Mistake}, \textit{Alignment}, and \textit{Pause}, and one control pattern \textit{one-dimension control}. 
We found that instead of being random and rare, \thingname{} frequently appear in non-expert demonstrations across all participants, and the distribution of \thingname{} is associated with the robot task temporarily and spatially. Moreover, we found that users' control patterns are affected by the control interface.
Our findings highlight the need for better models of sub-optimal demonstrations, offering insights to improve LfD algorithms and reduce gaps between lab-based training and real-world applications.
All the demonstrations, infrastructures, and annotations are available at \url{https://github.com/AABL-Lab/Human-Demonstration-Sidetracks}.

\end{abstract}
\section{Introduction}
Learning from human demonstration (LfD) has emerged as a crucial technique for robot learning \cite{ravichandar2020recent}, and has been successful in many contexts \cite{team2024octo}\cite{ jang2022bc}\cite{johns2021coarse}\cite{ padalkar2023open}, ranging from single-step to long-horizon tasks. 
One challenge in LfD is that high-quality demonstrations are hard to obtain \cite{yu2024much}. 
More recent work has demonstrated that using non-expert demonstrations could reduce the data needs.
However, there is little research on understanding and modeling non-expert demonstrations.
Much prior work assumes that non-expert demonstrations are low-quality and noisy\cite{beliaev2022imitation}, and mitigates the imperfections by smoothing the non-optimal behaviors out \cite{sasaki2021behavioral, yu2021active, yu2023thumbs, fang2025charm}.  
Our key insights are that: \textit{not all non-optimal behaviors from non-experts are random errors and randomly distributed: instead they appear in structured ways.}
Thus, in this work, we focus on identifying and investigating patterns of the noise in non-expert demonstrations. 

Modeling non-expert behaviors can be beneficial for both learning and validation. For instance, one important technique to enable learning from non-expert demonstrations is modeling non-expert demonstrations explicitly and designing learning algorithms to be able to adapt to these noisy or sub-optimal demonstrations \cite{robotics7020017, yuan2023goodbetterbestselfmotivated, chen2020learningsuboptimaldemonstrationselfsupervised}.
Another application is to help design better oracles. 
Oracles are widely used in prior work to provide non-expert demonstrations to validate algorithm performance \cite{9775674}, and non-expert behaviors are modeled by injecting noise into expert trajectories generated by machine-learning methods\cite{sasaki2020behavioral, laskey2017dartnoiseinjectionrobust}. 
Inaccurate modeling of non-expert behaviors can result in a performance gap between experiments and real-world scenarios.
Methods that work with oracle demonstrations might not perform well with real human demonstrations. 

However, few works have studied patterns in non-experts' behavior when giving demonstrations. Prior work often describes non-expert demonstrations using relatively simple patterns,  such as expert demonstrations with added noisy actions\cite{sasaki2020behavioral}, or expert policy mixing with random policies\cite{liu2022robust}, which fail to capture the diverse strategies or accommodation techniques non-experts use to succeed. 

\begin{figure}
    \centering
    \includegraphics[width=0.97\linewidth]{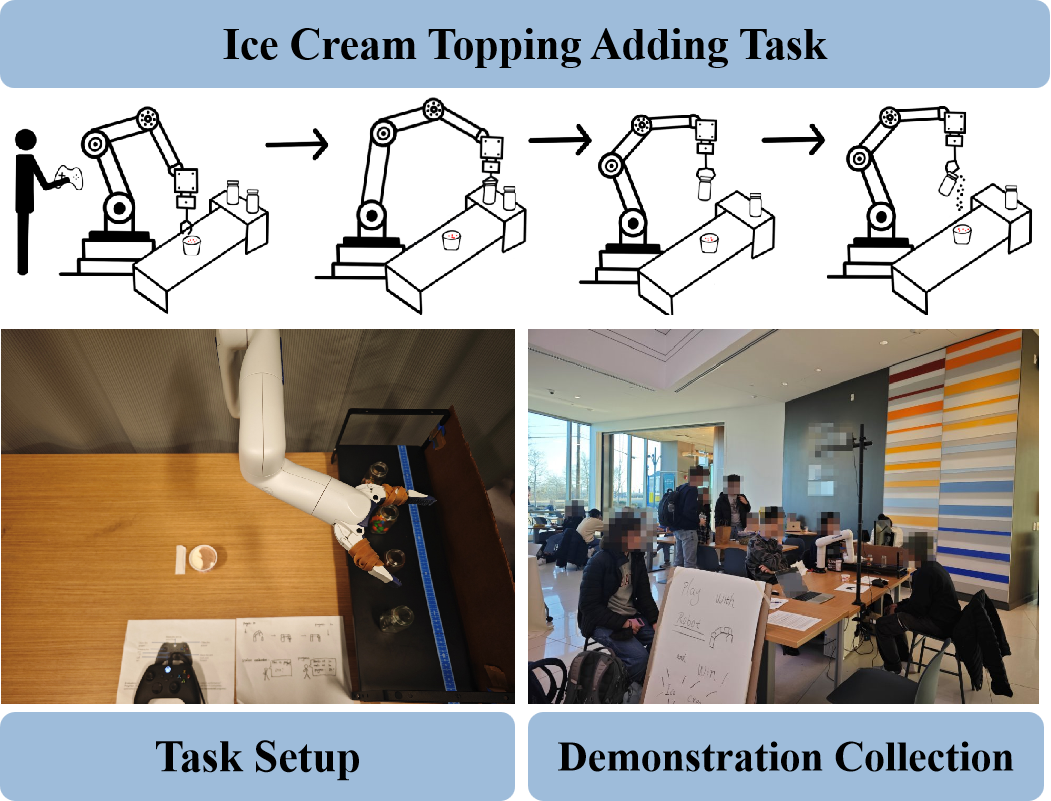}
    \caption{Public space study and ice cream topping adding task.
     Participants provided demonstrations in a non-lab setup with a long-horizon task. 
     The goal of the task is to control the robot arm to first pick up one of the four topping jars, and then pour the toppings onto an ice cream. 
    }
    \label{page1}
\end{figure}
Our goal is to investigate non-expert behaviors and explore the patterns of noise in non-expert demonstrations. 
In this work, we analyzed data from a public space study with 40 non-expert demonstrators with a long-horizon task, from our previous work initially introduced to test a novel feedback mechanism \cite{yu2024much}. 
We closely defined the frequent and patterned non-optimal behaviors that non-experts performed during demonstrating as \textit{\thingname{}}. 
To identify \thingname{} in the collected demonstrations,
 we designed a graphical interface and an open-coded book, annotated the demonstrations with two robot experts, and cross-validated the labels with a third expert. 
 All codes and data we annotated are published online at \href{https://github.com/AABL-Lab/Human-Demonstration-Sidetracks}{Demonstration-Sidetracks GitHub}.
We found that \thingname{} frequently exist in non-expert demonstrations across all demonstrators. We also showed that \thingname{} are not randomly distributed in demonstrations.
When and where \thingname{} appeared have temporal and spatial relationships with the task. 
The \thingname{} were more frequently identified when the sub-goal changed or when accurate manipulation or perceptions were required. 
Moreover, user behavior patterns are also associated with the control interface.

The main contributions of this work are characterizing and categorizing non-optimal behaviors in human demonstrations.
We found that non-optimal behaviors are beyond random errors, and possess spatial and temporal structures. 
To the best of our knowledge, this is the first work that systematically characterizes non-experts' non-optimal manipulation patterns when providing demonstrations.
Our work will be beneficial for future LfD algorithm development to better learn from noisy demonstrations and for oracle design to reduce the gap between in-lab datasets and real-world demonstrations. 

\section{Background}
A wide variety of learning from demonstration methods has emerged in the past few decades. 
Overcoming and understanding imperfections in human demonstration is crucial for the robustness of LfD methods, especially in real-world scenarios\cite{xu2022discriminator}.
\subsection{Learning from demonstration}
Learning from demonstration allows robots to learn a task policy from demonstrations provided by human teachers\cite{chernova2014robot}. 
Behavioral Cloning (BC) is a straightforward way of learning from demonstration\cite{pomerleau1988alvinn}. 
BC is capable of learning complex robot behaviors and performant general robot policies with sufficient and numerous demonstrations\cite{florence2022implicit}\cite{chi2023diffusion}\cite{team2024octo}. 
Inverse reinforcement learning (IRL) views the LfD process as recovering reward functions from human demonstrations and learning a robot policy via reinforcement learning from inferred reward functions\cite{abbeel2004apprenticeship}\cite{ziebart2008maximum}.
Other methods drawing intuition from IRL inherit its basic idea but vary in terms of network structures. \cite{ho2016generative}\cite{fu2017learning}
Most regular LfD methods require high-quality demonstrations to work\cite{xu2022discriminator}.
Difficulty in obtaining sufficient expert demonstrations forms a major challenge for LfD methods. As a result, efficiently learning from non-expert demonstrations is crucial to the wide deployment of LfD-based algorithms. 
This addresses the need to look into actual patterns in non-expert demonstrations. 

\subsection{Learning from non-expert demonstrations}

Enabling LfD to learn from sub-optimal or non-expert demonstrations is a promising approach to lifting data availability and robustness for robot learning 
\cite{wang2023identifying, gopalan2022negative}. 
Learning methods need to infer the distribution of the optimal policy from noisy or sub-optimal distributions.
    One applicable method is by introducing extra human knowledge, such as ranking\cite{brown2019extrapolating}, preference\cite{kuhar2023preference}, feedback \cite{10.1145/3610977.3634925}, and confidence\cite{zhang2021confidence,wu2019imitation}. 
Other methods introduce prior knowledge about the possible distribution of optimal policies. This prior knowledge can come from a small set of experts or optimal demonstrations\cite{xu2022discriminator}\cite{yu2023offline}, or pre-designed metrics about the task \cite{bu2024aligning}. Prior knowledge allows LfD algorithms to implicitly rank or weight sub-optimal demonstrations based on their distance to given optimal demonstrations or desired metric values, resulting in a better inference about the true optimal policy. However, including extra human knowledge can be laborious and time-consuming. Also, this extra information is typically task-specific so it can not be used ``off-the-shelf'' to enable learning from non-expert demonstrations in new scenarios.

\subsection{Modeling Noisy Demonstrations}
In order to learn from non-expert demonstrations without introducing extra human feedback or prior knowledge, modeling sub-optimality becomes necessary for learning from sub-optimal demonstrations. 
The modeling of sub-optimality includes the modeling of the type of distribution\cite{beliaev2024inverse}, scale\cite{sasaki2020behavioral}, and their relationships to other variables\cite{beliaev2022imitation}. 
With the modeling of sub-optimal demonstrations, optimal policies can be recovered through selective matching\cite{sasaki2020behavioral}, denoising \cite{yuan2023good}, automatic ranking and weighting \cite{beliaev2022imitation}.
Despite increasing focus on learning from imperfect human demonstrations, work that uses real-world imperfect human demonstrations is still relatively rare. Prior work generally has relatively simple assumptions on sub-optimality modeling: like being a mixed Gaussian distribution of optimal and noise \cite{liu2022robust}\cite{beliaev2024inverse}\cite{beliaev2022imitation} or having optimal the part being dominant while sub-optimality only takes up a small part \cite{sasaki2020behavioral}.

Our work differs from prior work by recognizing that non-optimal behaviors are natural for human manipulation, manifesting in frequently distributing in human demonstrations with spatial and temporal patterns. 
Thus, our work focuses on categorizing and identifying patterns in non-optimal demonstrations.
The demonstrations we used in this work were collected outside the lab with random people. Our data and results can be expected to better reflect robot-in-the-wild deployments than in-lab setups. 
\begin{figure*}
    \centering
    \includegraphics[width=0.8\textwidth]{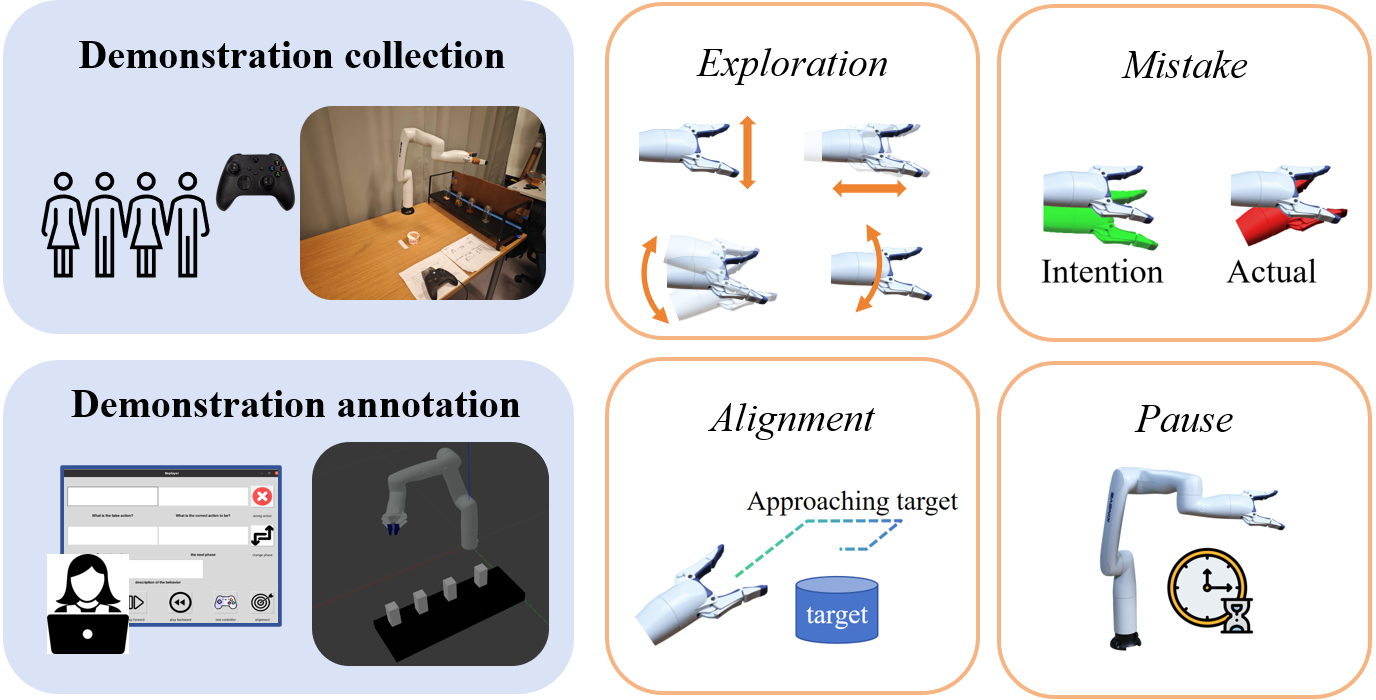}
    \caption{Experiment pipeline and demonstration sidetracks. Non-expert demonstrations were collected by having participants control a real robot. We then replayed demonstrations in simulation to annotate the \thingname{}. We identified four types of \thingname{} and one control pattern. We illustrated \thingname{} -- \textit{Exploration}, \textit{Mistake}, \textit{Alignment}, and \textit{Pause} -- on the right side of our figure.
    }
    \label{pipeline}
\end{figure*}

\section{Methodology}
Providing demonstrations to a robot task can be complex and highly demanding. 
Unlike well-trained agents, 
humans exhibit sub-optimal behaviors while manipulating robots. 
However, not all imperfect behaviors are meaningless.  We believe that some non-optimal behaviors are beyond noise or mistakes, they are actually performed by the demonstrators intentionally or subconsciously to provide successful demonstrations. 
Additionally, human errors may also not appear randomly in the demonstrations. Those errors might be spatially or temporally associated with the task. 
In this work, we conducted an exploratory analysis to investigate noisy behaviors in non-expert human demonstrations.  
\textbf{Our key insight} is that:
\begin{enumerate}[label={}]
  \item \textit{Non-optimal behaviors in human demonstrations are structured with patterns.}
\end{enumerate}
The demonstrations came from 40 non-expert demonstrators who were asked to provide demonstrations over a long horizon task.
In this work, we annotated all the demonstrations using an open-coded book with a graphical interface.

\subsection{\ThingName{}}
Prior work interprets non-optimal behaviors in human demonstrations as random manipulation noise or incorrect behaviors with underlying optimal control policies \cite{beliaev2024inverse}\cite{beliaev2022imitation}\cite{liu2022robust}. 
However, unlike trained agents or oracles, which can control all action dimensions instantly and are either provided with accurate observation or trained to reduce perception error, humans are limited by the control interface, reaction time, and perception accuracy.  

Intuitively, in order to complete the task while maintaining a reasonable mental workload, humans accommodate these limitations by controlling a limited degree of freedom at a time, performing redundant but harmless behaviors, or spending more time while giving demonstrations. 
These accommodation behaviors would appear consistently both in the user's overall policy and specific control behaviors, and would have special task-related temporal and spatial characteristics.  
As a result, human demonstrations tend to be different from machine demonstrations by containing not only noisy actions but also accommodation behaviors. 
In this work, we define non-optimal behaviors in human demonstrations as \textit{\ThingName{}}. 

\subsection{Experiment Setup}
The demonstration data analyzed in this work was collected as part of a validation study for a novel teaching signal \cite{yu2024much}.  We review the study design here, which is detailed described in \cite{yu2024much}.
The robot used for the data collection was a Kinova Gen3 Lite robot with six degrees of freedom. Participants used an Xbox controller to control the robot arm. 
The task used to collect human demonstrations is an ice cream topping task. 

The setup and the control interface are shown in \autoref{pipeline}.
The task consisted of six sub-tasks: move down to jar, approach jar, grasp jar, lift jar, move to ice cream cup, approach ice cream cup, and pour.
Participants first control the robot to reach a jar they selected on a shelf, pick up the jar,  navigate to the ice cream location, and then pour the toppings into the ice cream cup. 
This task is suitable for our study because it features both simplicity and diversity of primal sub-tasks. The overall task is relatively simple and has an appropriate length for manual annotation. It includes diverse sub-tasks and primal actions that cover some of the most common robot skills, including free-space reaching, approaching small objects, grasping, aligning the robot to the target position, and object pose manipulation\cite{padalkar2023open}. 
\begin{figure*}
    \centering
    \includegraphics{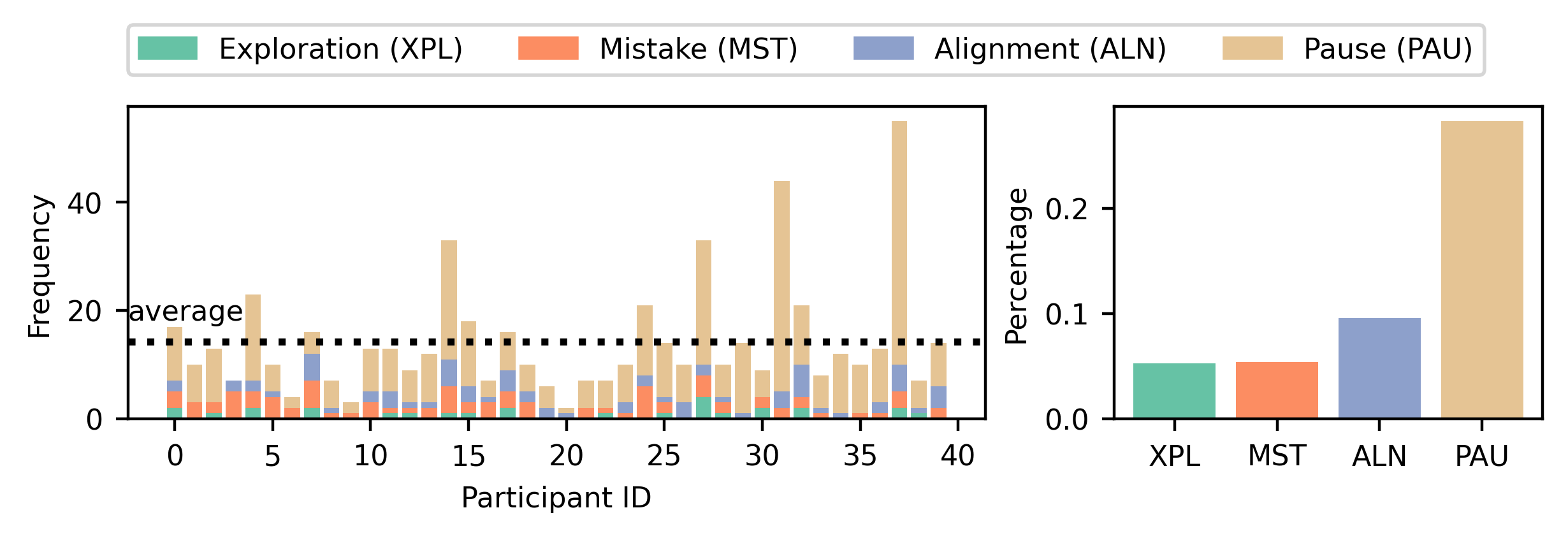}
    \caption{ Demonstration sidetracks frequency and ratio. 
    The left figure shows the number of \thingname{} observed across all demonstrators. The X-axis is the participant ID, and the Y-axis is the number of each type of demonstration sidetrack. The figure on the right side shows the percentage of time spent on each type of \thingname{} relative to the total time.
    The results indicate that \thingname{} widely and frequently exist in non-expert demonstrations. 
    }
    \label{fig:frequency per p}
\end{figure*}

All control actions were in end-effector space, and actual joint motions were calculated in real-time using forward kinematics.
The end-effector movements were decomposed into several primitive movements, including translations along the X, Y, and Z axes, rotations in roll, pitch, and yaw, as well as opening and closing of the gripper.

\subsection{Experiment procedure}
The public space study was held in the atrium of a university building to collect non-expert demonstrations in a non-lab setup, as shown in \autoref{page1}.
The study was approved by Tufts University's Institutional Review Board (IRB).
Participants were recruited from random people walking past the experiment setup. 
40 participants were recruited in total. 
22 were male, 14 were female, and 4 preferred not to say.
Participants were asked to provide one demonstration and had up to three minutes to practice the task. 
Participants only had one shot to demonstrate the task, and could not retry if they failed. 
The demonstration data were recorded at a frequency of 5Hz in the form of joint positions.

\subsection{Open-Coded Book}
We focus on the following four types of \thingname{} in this work: 
\subsubsection {Exploration} A sequence of primal movements that include multiple dimensions of freedom, and often happens in the begin of the task and are irrelative to completing the task. 
\subsubsection{Mistake} A sequence of homogenous primal movements that has no contributions to, or harms the completion of the task.
\subsubsection{Pause} Significant pause during the execution of the task. In our case, one second with no control inputs is considered a pause. 
\subsubsection{Alignment} Back-and-forth movements that align the robot with the target object.
We used one additional features to help to identify the intention of alignment, which is actual precision increase, meaning the robot is closer to the target after this behavior. 

Here, we define a primal movement as a movement that is mapped to a single input source on the controller interface, i.e., one button or joystick. In the case of our study, the seven primal movements are moving along the X, Y, and Z axes, roll, pitch, yaw, and operating grippers. 

In addition to \thingname{}, we also focus on one control pattern:
\subsubsection{One-dimension control} Tendency of control using a single primal movement at a time.

\subsection{Replay and annotation}

We recreated our experiment setup in a simulation environment, and annotated the demonstrations in the simulation. 
Simulation allows us to control the speed of replaying and pause or play backward at any step, which helps us to catch more subtle non-expert behavior and annotate their precise starting and ending times.
All \textit{pauses} were automatically annotated by a script. 
In addition to \thingname{},
we also annotated task phases to study the temporal relationships between non-expert behaviors and task phases.

We designed a graphical user interface to help with data annotation. 
The interface includes three buttons for controlling the replay,
three buttons for labeling \thingname{}, one button for marking phase change, and five input boxes for additional comments. 
This interface allows annotators to pause the replay, play forward, play backward, label behaviors and task phases, and add additional descriptions, such as start and end steps for the labels. 

Two robot experts conducted the annotation, and the annotations were crosschecked by a third robot expert. Behaviors that appear both in the original annotations and crosscheck annotations are adopted for our data analysis. 

\begin{figure*}
    \centering
    \includegraphics{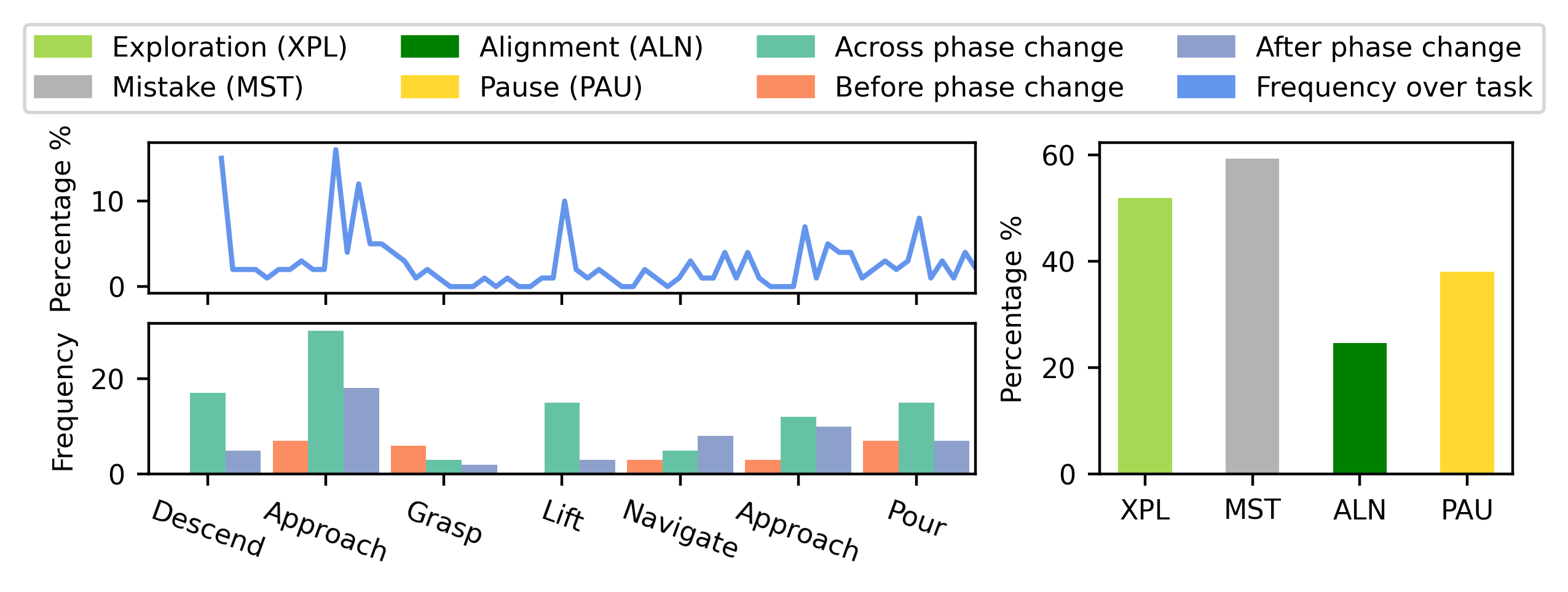}
    \caption{Temporal relationships between task phases and \thingname{}. 
    The upper-left part shows the percentage of \thingname{} in different sub-tasks. The bottom-left part shows the number of \thingname{} happened within 40\% percent timesteps window around the phase change. The X-axis displays different phases of the task in order, and the Y-axises are percentages and frequency separately. 
    The right part shows the percentage of each type of \thingname{} happening within a 4-second window around the task phase change.
    We found that the occurrence of \thingname{} is associated with the change of the task phases. 
    }
    \label{fig:temporal}
\end{figure*}

\section{Results}

We manually annotated exploratory actions by replaying 40 real-world robot demonstrations from non-expert demonstrators in simulation according to our proposed open-coded book.
The main goal for conducting data analysis was to identify \thingname{} and unveil the underlying patterns of these behaviors. 
We found that \thingname{} widely existed in the demonstrations,  \thingname{} more frequently appeared around the target objects or when the sub-task has changed, and demonstrators tended to control one dimension at a time. 

\subsection{Demonstrations Sidetracks are frequent in demonstrations}

In this subsection, we demonstrate the importance of identifying and characterizing \thingname{} by showing that these \thingname{} widely existed in human demonstrations, even in demonstrations that successfully complete the task. 

We show the amount of \thingname{} for each participant in \autoref{fig:frequency per p}. 
We found that \thingname{} widely exist across all demonstrations: while 33 out of 40 non-expert participants successfully demonstrated the task, the average number of \thingname{} is $14.48 \pm 10.54$  per participant. 
We calculated the proportion of the timesteps associated with \thingname{} relative to total recorded timesteps.  
Demonstration sidetracks comprised about 48\% of the overall task's recorded timesteps. This suggests that, contrary to the common assumption that imperfections only take up a small part in non-expert demonstration, \thingname{} are frequent and take up a large portion of the overall task demonstration.
Additionally, we observed a high diversity in \thingname{} types in a single demonstration, with 62.5\% of the demonstrations possessing three or more types of \thingname{}.

\subsection{\ThingName{} are more likely to happen when sub-task changes}


One key assumption when using simple distributions to sample noise is that noise is independent from the task itself. However, we show that this assumption does not hold for actual human demonstrations by showing the relevance between \thingname{} and the task's progress. 

We analyzed the relative time when \thingname{} occurred during different task phases, which are descending to the height of the jar, approaching the jar, grasping the jar, lifting, navigating to the ice cream, approaching the ice cream, and pouring the toppings. The results are shown in \autoref{fig:temporal}. We found that \thingname{} were more likely to occur near the moment when the task phase changed. 54.7\% of overall \thingname{} behaviors occurred in the 40\% range across the instance of the task change. This is likely due to the different sub-goals of different task phases, along with a change in the control pattern for participants, resulting in increasing \thingname{}. 
Additionally, we analyzed the absolute time when \thingname{} happened. Specifically, we analyzed the percentage of different \thingname{} happening within four seconds from the instance of the task phase change. Results show that 52\% of \textit{Exploration}, 59\% of \textit{Mistake}, and 38\% of \textit{Pause} happened within four seconds from task-phase-change instances, while the average time length of a demonstration is about 84 seconds. 
This further reveals that humans tend to have \thingname{} near task phase change temporally.

\begin{figure}
    \centering
    \includegraphics[scale=0.5]{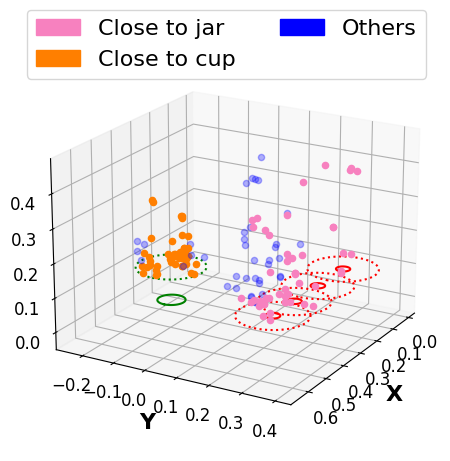}
    \caption{The spatial distribution of \textit{Alignment} behaviors. The points represent the position of the robot end-effector positions at the start and end of each \textit{Alignment} behavior.
    The red circles represent the positions of four jars, while the green circle represents the ice cream cup. 
    Points within 0.1 meters from the jars and the cup are highlighted in orange and pink. Other points outside this range are illustrated in blue.
    We found that \textit{Alignment} behaviors more frequently occur around target objects. 
    }
    \label{online2}
\end{figure}
\subsection{\textit{Alignment} happens more when accuracy is required for controlling robot}

In this subsection, we examine the spatial distribution of \thingname{}. 
Simulated noise is usually sampled at random steps and rarely considers the actual robot and target locations. 
We demonstrate that object positions are an important task-related factor affecting human non-optimal behavior by showing how precision requirements affected the spatial distribution of \thingname{}.

We analyzed the relationship between the number of \thingname{} and object locations. We showed that \textit{Alignment} occurred more often when the task was to approach a certain target position, as shown in \autoref{online2}. 73\% of all \textit{Alignment} behaviors started and ended within 0.1-meter range from the target. Among these, 40\% \textit{Alignment} behaviors started and ended near the jars, and 33\% of all alignment behaviors happened near the ice cream cup, which was the target for the pouring task.
This suggests that \thingname{}, especially \textit{Alignment}, are more likely to happen when approaching a target object. This is possibly due to the requirement for control accuracy, which resulted in more alignment to get to the desired positions and more pauses for reaction.

\begin{figure}
    \centering
    \includegraphics[width=0.95\linewidth]{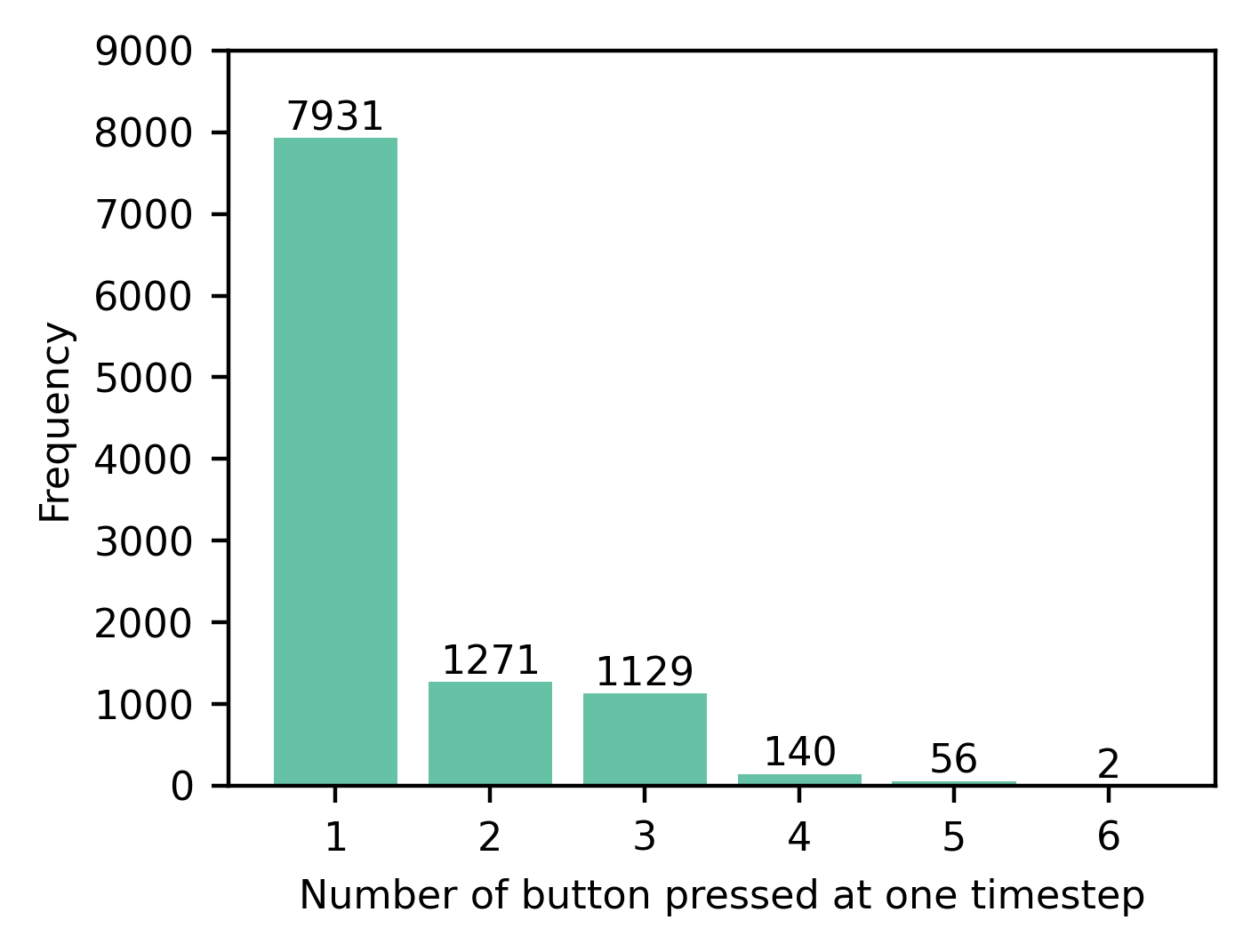}
    \caption{ 
    The number of dimensions controlled by participants at a single timestep. The X-axis shows the number of dimensions controlled at a single recorded timestep, while the Y-axis shows the frequency.
    Results show that, for 75\% of the timesteps,  participants only controlled one dimension, i.e., one button or one joy stick was pushed. 
    }
    \label{fig:num-buttons}
\end{figure}

\subsection{Effect of Control Interface}
In addition to the imperfect manipulations that led to changes in control patterns, we also found that the control interface affected participants' overall manipulation strategies. 
We found that there is a common tendency in participants to control only one dimension at a time in their manipulation patterns.
We analyzed the number of buttons pushed during each record timestep by measuring the position change of the robot end effector before and after each recording timestep and extracted the number of actions taken. The results showed that when the robot was moving, 75\% of all timesteps only recorded a single action, whereas 12\% recorded two actions and 10\% recorded three actions, as shown in \autoref{fig:num-buttons}. This tendency of controlling fewer dimensions simultaneously can cause human policies to deviate significantly from the machine-generated policies.

\section{Discussion}
We found that human imperfections when giving demonstrations \textbf{ are not just simple noise or randomly distributed}. Rather, these imperfections usually have distinct patterns, are often intentional, and are associated with the control interface, which can cause real-world human demonstrations to differ from machine-generated demonstrations. 
We defined these meaningful or patterned imperfections as \thingname{}.

We believe the differences between machine-generated demonstrations and human demonstrations lie in the following factors. 
\textbf{Human control of robots is limited by the input interface.} For instance, in our experiment, participants controlled the robot using an Xbox controller, and each button or joystick controlled one dimension of freedom.
Controlling movements in multiple dimensions requires a higher mental load, and thus was less preferred even if it was more optimal. 
\textbf{Humans need reaction time when controlling robots}, which could result in frequent pauses.
Lastly, unlike well-trained agents or oracles, which are provided with unambiguous observations and trained to have consistent reactions, 
\textbf{human perceptions and controls are less precise}.
To succeed in the task, demonstrators may perform a sequence of sub-optimal actions instead of one optimal action to mitigate errors. 


One limitation of this work is that we only used demonstrations from one task and one robot.
Even though the task has covered a good range of sub-tasks (e.g., picking, grasping, and pouring), the work will benefit from including non-manipulation tasks.
For future work, we will develop quantitative metrics for \thingname{} and methods that could identify \thingname{} automatically. 
Another future work is to design a method that injects \thingname{} into synthetic machine-generated demonstrations to generate more realistic demonstrations for improving LfD algorithm development and performance validation. 

\section{Conclusion}

In this work, we explored systematic human non-optimal behaviors, defined as \textit{\ThingName{}}, in a real-world robot demonstration task.  We identified four categories of \thingname{} and one control pattern from human demonstrations. 
We showed that \thingname{} are frequent, associate with the task both temporally and spatially, 
and  are affected by human control policies. 
Our findings highlight the need for a realistic modeling of sub-optimality in human demonstrations while developing learning from demonstration methods. We believe this work lays a foundation for bridging the gap between laboratory-based setups and real-world scenarios.

%
%
\bibliographystyle{IEEEtran}
\bibliography{main}
\end{document}